\documentclass[letterpaper, 10 pt, conference]{ieeeconf}  

\IEEEoverridecommandlockouts                              
\let\proof\relax

\usepackage{dsfont}
\usepackage{amsthm}
\usepackage{amsmath}
\usepackage{amssymb,amsfonts}
\usepackage{dsfont}
\usepackage{bbm}
\usepackage{paralist}
\usepackage{xcolor}
\usepackage{mathtools}

\usepackage{tikz}
\usetikzlibrary{chains,automata,positioning,arrows}
\tikzset{every loop/.style={in=-120,out=-60,looseness=4}}

\usepackage{cite}

\newtheorem{theorem}{Theorem}

\newtheorem{definition}{Definition}

\newtheorem{proposition}{Proposition}
\newtheorem{corollary}{Corollary}

\def\EE{{\mathbb E}}
\def\PP{{\mathbb P}}
\def\RR{{\mathbb R}}

\def\kl{\mathrm{KL}}

\newcommand{\ep}{\hfill $\Box$}

\newcommand{\argmax}{\arg\!\max}
\newcommand{\set}[1]{\mathcal{#1}}

\title{\LARGE \bf
Regret Analysis in Deterministic Reinforcement Learning
}

\author{Damianos Tranos and Alexandre Proutiere
\thanks{This work was supported by the Wallenberg AI, Autonomous Systems and Software Program (WASP) funded by the Knut and Alice Wallenberg Foundation.}
\thanks{D. Tranos and A. Proutiere are with the Division of Decision and Control Systems, School of Electrical Engineering and Computer Science, Royal Institute of Technology (KTH), Stockholm, Sweden. Emails: 
        \{{\it tranos@kth.se, alepro@kth.se }\}.}%
}

\begin{document}

\maketitle
\thispagestyle{empty}
\pagestyle{empty}

\begin{abstract}

We consider Markov Decision Processes (MDPs) with deterministic transitions and study the problem of regret minimization, which is central to the analysis and design of optimal learning algorithms. We present logarithmic problem-specific regret lower bounds that explicitly depend on the system parameter (in contrast to previous minimax approaches) and thus, truly quantify the fundamental limit of performance achievable by any learning algorithm. Deterministic MDPs can be interpreted as graphs and analyzed in terms of their cycles, a fact which we leverage in order to identify a class of deterministic MDPs whose regret lower bound can be determined numerically. We further exemplify this result on a deterministic line search problem, and a deterministic MDP with state-dependent rewards, whose regret lower bounds we can state explicitly. These bounds share similarities with the known problem-specific bound of the multi-armed bandit problem and suggest that navigation on a deterministic MDP need not have an effect on the performance of a learning algorithm.
\end{abstract}

\section{INTRODUCTION}

Reinforcement Learning addresses the optimal control problem of an unknown dynamical system, which is traditionally modeled as a Markov Decision Process (MDP). The stochastic nature of this problem induces an exploration-exploitation dilemma where the decision-maker must balance between decisions which give insight into the dynamics of the system and decisions which, given the information available, are considered optimal. This dilemma is quantified mathematically by the notion of regret, defined as the difference between the cumulative reward obtained by the decision maker, and that obtained by an oracle who always makes the best decisions.

The design of an optimal algorithm is then equivalent to the design of an algorithm that minimizes regret. A natural question then is, what is the minimum regret that can be achieved by a learning algorithm, given a certain class of MDPs. The answer to this question is expressed in the form of \emph{problem-specific} regret lower bounds. By problem-specific, we mean that there is an explicit dependence on the system parameter. In contrast with minimax bounds which are conservative in nature, these bounds accurately quantify the fundamental performance limits attainable by a learning algorithm. Moreover, they are instrumental in the design of optimal learning algorithms, as has been evidenced by \cite{cappe2013kullback}, \cite{magureanu2014lipschitz}, and \cite{ok2018exploration}.

In this work we study the fundamental performance limits for a class of MDPs with finite state-action spaces and deterministic transitions. The reward obtained with each transition is stochastic and must be learned by the decision-maker. Knowledge of the transitions can be interpreted as knowledge of the structure of the problem, and so we are inspired by \cite{lakshmanan2015improved} and \cite{ok2018exploration} who quantify the possible performance gains when the structure of their respective problems is exploited. 

\subsection{Related Work}
The first asymptotic, logarithmic, and problem-specific regret lower bound for MDPs was provided, for the case of known rewards, by \cite{burnetas1996optimal} who also devised an algorithm whose regret upper bound asymptotically matches their lower bound. This lower bound is extended by \cite{ok2018exploration} to the case of unknown rewards as well as the case of MDPs with structure. For both cases, \cite{ok2018exploration} propose an algorithm with matching regret upper bounds.

The case of communicating MDPs is addressed by the works of \cite{auer2007logarithmic}, \cite{jaksch2010near}, \cite{bartlett2009regal}, and \cite{filippi2010optimism}. They provide logarithmic and finite-time regret guarantees, but at the expense of optimality, i.e., a much larger constant in front of $\log T$.

To the best of our knowledge, regret minimization for deterministic MDPs has only been addressed by \cite{ortner2010online}. Therein, they propose UCYCLE, an extension of the UCB1 algorithm \cite{auer2002finite} which selects optimistically among cycles of state-action pairs. While the algorithm is claimed to achieve logarithmic regret, it does not match their proposed minimax lower bound and the analysis of the algorithm is based on a relaxed notion of regret.

\subsection{Contributions}
In this work, we present a problem-specific asymptotic regret lower bound valid for any MDP with deterministic transitions and any learning algorithm.

By relating deterministic MDPs to graphs and analyzing them in terms of their cycles, we identify a class of deterministic MDPs whose cycles are disjoint. These problems possess a decoupling property which allows us to simplify our lower bound considerably from an infinite-dimensional optimization problem, to two nested finite-dimensional ones, making it possible to evaluate it numerically.

We exemplify our simplified bound on two specific problems, that of deterministic line search and that of deterministic MDPs with state-dependent rewards. For both problems, we are able to solve the optimization problems associated with their lower bounds analytically and thus state them explicitly. These bounds are analogous to the lower bound derived by Lai and Robbins \cite{lai1985asymptotically} for the multi-armed bandit problem, a fact which suggests that they are tight. This also indicates that an optimal algorithm need not experience additional regret resulting from navigating a deterministic MDP.

\section{PROBLEM FORMULATION}

In this section we introduce Deterministic Markov Decision Processes and define precisely the notion of regret. We then present relevant graph-theoretic notation which we will use in our study of regret lower bounds.

\subsection{Deterministic Markov Decision Processes}

We denote by $\Phi_D$ the set of all Deterministic Markov Decision Processes (DMDPs), which consists of discrete-time controlled Markov Chains with deterministic transitions. A DMDP is defined as the quadruple $\phi = (\set{S}, \set{A}, p_\phi, q_\phi)$, where $\set{S}$ is the finite state space and $\set{A}$ is a finite set of actions. They have respective cardinalities $S$ and $A$. Given a state $s$ and an action $a$, a transition to the next state $s'$ occurs with probability $p_\phi(s'|s,a) = 1$. A random reward is also sampled from a one-parameter exponential distribution $q_\phi(\cdot|s,a)$, parameterized by its mean $r_\phi(s,a)$. A decision maker chooses actions according to a policy $\pi$, defined as a distribution over $\set{A}$ based on the history of states, actions, and rewards. We denote by $\Pi$ the set of all policies. A notable subset, is the set of all stationary, deterministic, Markov policies $\Pi_D$, which is the set of mappings $\pi :\set{S} \to \set{A}$. 
 
Given a DMDP, the expected cumulative reward up to time step $T$ obtained by the decision maker who follows a \emph{policy} $\pi \in \Pi$ from the initial state $s$ is $V_T^\pi(s) = \EE_{s}^\pi \left[\sum_{t=1}^{T} r_\phi(s_t,a_t) \right]$. Here, $\EE_{s}^\pi\left[ \cdot \right]$ is the expectation under policy $\pi$ given that $s_1 = s$. 
The objective of the decision-maker, when faced with an unknown DMDP $\phi$, is to devise a policy $\pi\in \Pi$ that maximizes $V_T^\pi(s)$ or, equivalently, minimizes the \emph{regret} up to time $T$: $R_T^\pi(s) := V^\star_T(s) - V^\pi_T(s)$  where $V^\star_T(s) := \sup_{ \pi \in \Pi} V^\pi_T (s)$.

A DMDP $\phi$ is \emph{communicating} if for any pair of states $s,s'\in \set{S}$ there exists $\pi\in \Pi_D$ such that $s'$ is accessible from $s$.
For any communicating $\phi$ and any policy $\pi\in \Pi_D$, we denote by $g^\pi_\phi(s)$ the \emph{gain} of $\pi$ with initial state $s$: $g_\phi^\pi (s):=\lim_{T\to\infty}{\frac{1}{T}}V_T^\pi(s)$, which is known to always exist in this problem setting (see Proposition 8.1.1 in \cite{puterman2014markov}). We denote by $\Pi^\star(\phi)$ the set of \emph{gain-optimal} (or optimal) policies, i.e., the stationary policies with maximal gain: $\Pi^\star(\phi):=\{ \pi \in \Pi_{D}: g_\phi^\pi (s)= g_\phi^\star(s)~\forall s \in \set{S}\}$, where $g_\phi^\star (s):=\max_{\pi \in \Pi} g_\phi^\pi(s)$. It is known (see Theorem 8.3.2 in \cite{puterman2014markov}), that if $\phi$ is communicating then the maximal gain, denoted by $g^\star_\phi$, is constant.
\subsection{Graphs and Cycles}

We associate a DMDP with a graph $G = (V,E)$ where the states are the vertices, $V = \set{S}$, and the state-action pairs are the edges, $E = \set{S}\times \set{A}$. A walk is defined as a tuple of edges $((s_1,a_1),\dots,(s_p,a_p))$ such that $p_\phi(s_{i+1}|s_i,a_i) = 1$ for $1 \leq i < p$. Given two walks $U = (u_1,\dots,u_p)$ and $W = (w_1,\dots,w_q)$, we define their concatenation $U \cdot W = (u_1,\dots,u_p,w_1,\dots,w_q)$. A path is a walk without repeated vertices. 

A walk or path is closed if the its final edge connects to its initial vertex, i.e., $p_\phi(s_1|s_p,a_p) = 1$. A closed walk is referred to as a \emph{cycle}, while a closed path is referred to as a \emph{simple cycle}. Since the DMDPs we consider are finite, there exists a finite number of simple cycles in any given DMDP $\phi$ (up to a permutation). We denote the set of all such simple cycles by $\set{C}$. 

In a DMDP $\phi$ with initial state $s$, every stationary deterministic policy $\pi \in \Pi_D$ induces a cycle $C^\pi_\phi(s)$. The gain of this policy can be written as:
\begin{align*}
g_\phi^\pi(s) = \frac{1}{|C^\pi_\phi(s)|}\sum_{(s,a)\in C^\pi_\phi(s)}r_\phi(s,a).
\end{align*}
Since the gain of a gain-optimal policy does not depend on the initial state, it follows that for any $\pi \in \Pi^\star(\phi)$ there exists an optimal cycle $C$ such that:
\begin{align*}
g_\phi^\star = \frac{1}{|C|}\sum_{(s,a)\in C}r_\phi(s,a):= g_\phi(C).
\end{align*}
More importantly, if such an optimal cycle exists, then there also exists an optimal simple cycle with the same gain:

\begin{proposition} \label{prop:optimal_circuit}
For any DMDP $\phi \in \Phi_D$, if there exists a gain-optimal policy $\pi$ that induces an optimal cycle $C$, then there also exists a gain-optimal policy $\pi'$ that induces an optimal simple cycle $C'$.
\end{proposition}

\noindent
\proof The proof consists of two parts. Firstly, we show that any cycle $C$ can be expressed as the concatenation of simple cycles $C_i$, up to a permutation. Secondly, we show that if $C$ is gain-optimal, then one of its components $C_i$ must also be gain-optimal.

As shown in \cite{komusiewicz2015sound}, we can express any walk $C$ as the concatenation of possibly empty paths $W_i$ and simple cycles $C_i\in \set{C}$, i.e, $C = C_1\cdot W_1\dots C_q\cdot W_q$. We define by $C^1:= W_1 \cdot W_2 \dots W_q$ the resulting sequence of edges after we have removed every simple cycle $C_i$. Then $C^1$ is also a cycle and thus can also be decomposed into simple cycles and possibly empty paths. We recursively apply this decomposition and removal step on each cycle $C^i$ such that iteration $C^N$ is empty. We then have $C = C^1\cdot C^2\dots C^{N-1}$ (up to a permutation), so that $C = m_1C_1\cdot m_2C_2\dots m_nC_n$, where $m_i$ is the number of times the simple cycle $C_i$ appears in cycle $C$, and $n$ is the cardinality of $\set{C}$. We thus have:
\begin{align*}
\sum_{(s,a)\in C} r_\phi(s,a) = \sum_{i=1}^{n} m_i\sum_{(s,a)\in C_i}r_\phi(s,a),
\end{align*}
and
\begin{align*}
|C| = \sum_{i=1}^{n}m_{i}|C_i|,
\end{align*}
which, letting $\bar{\set{C}}:= \{C_i\in \set{C}: m_i \neq 0\}$, leads to:
\begin{align*}
g_\phi(C) = \sum_{i=1}^{n}\frac{m_i|C_i|}{|C|}g_\phi(C_i) \leq \max_{C_i\in \bar{\set{C}}}g_\phi(C_i). 
\end{align*}
However, since $g_\phi(C)$ is the optimal gain, we must have that $g_\phi(C) \geq g_\phi(C_i)$, for all $C_i\in \set{C}$. Therefore, the following equality holds:
\begin{align*}
g_\phi(C) = \max_{C_i\in \bar{\set{C}}}g_\phi(C_i).
\end{align*}
Thus, for a gain-optimal policy $\pi$ which induces cycle $C$, there exists a simple cycle $C' = \argmax_{C_i\in \set{C}}g_\phi(C_i)$, induced by a policy $\pi'$, which is also gain-optimal.
\ep

As a consequence of this proposition, we will be restricting our attention to simple cycles and will henceforth abuse the notation by referring to them as cycles.


\section{REGRET LOWER BOUND FOR DETERMINISTIC MARKOV DECISION PROCESSES}

In this section, we state our regret lower bound valid for any communicating DMDP. To this aim, we introduce the following definitions and notations:

\begin{definition}
A policy $\pi\in \Pi$ is uniformly good if for all $\phi\in \Phi_D$, $s_1 \in \set{S}$, $\alpha>0$, and $(s,a) \notin C^\star_\phi$, we have $\EE_{\phi|s_1}^\pi[N_T(s,a)] = o(T^\alpha)$.
\end{definition}

For $\phi, \psi \in \Phi_{D}$ we denote by $\kl_{\phi \mid \psi}(s,a)$ the Kullback-Leibler divergence between the reward distributions $q_\phi$ and $q_\psi$ at state $s$ when action $a$ is chosen:
\begin{align*}
\kl_{\phi \mid \psi}(s,a) 
=
\int_{0}^1 q_\phi(r | s,a) \log \frac{q_\phi(r | s,a)}{q_\psi(r | s,a) }\lambda(dr).
\end{align*}

\begin{definition}

We define by $\Delta(\phi)$ the set of confusing models for DMDP $\phi$. Specifically, $\psi$ is \emph{confusing} if it is absolutely continuous with respect to $\phi$, $(i)$ has the same reward distributions at every state-action pair that is part of the optimal cycle of $\phi$ and $(ii)$ has optimal policies that are not optimal under $\phi$, i.e.,
\begin{align*}
\Delta(\phi) = \Big\{&\psi \in \Phi_D :  \phi \ll \psi, \\
&(i) \ \Pi^\star(\phi) \cap \Pi^\star(\psi) = \emptyset, \\
&(ii)\ \kl_{\phi \mid \psi}(s,a) = 0,~  \forall (s,a) \in C^\star_\phi \Big\}.
\end{align*}
\end{definition}

Finally, we denote by $N_T(s)$ and $N_T(s,a)$, the respective number of times, up to time $T$, that the state $s$ and state-action pair $(s,a)$ have been visited.

\begin{theorem} \label{thm:g-r-lower-det}
Let $\pi \in \Pi$ be a uniformly good policy. For all $\phi \in \Phi_D$ and initial $s_1 \in \set{S}$ we have:
\begin{align} \label{eq:g-r-lower-det}
\liminf_{T \to \infty} \frac{R^\pi_T (s_1,\phi)}{\log T} \ge C(\phi), 
\end{align}
where $C(\phi)$ is the value of the optimization problem:
\begin{align}
\min_{\eta \in \set{I}(\phi)\cap \set{N}(\phi)} \sum_{(s,a)\in \set{S}\times \set{A}} \eta(s,a)(g^\star_\phi - r_\phi(s,a)), \label{eq:g-r-obj-det}
\end{align}
with information constraints:
\begin{align}
\set{I}(\phi):= \Big\{&\eta \in \RR_{+}^{S \times A}: \forall \psi \in \Delta(\phi), \nonumber \\
&\sum_{(s,a) \notin C_\phi^\star} \eta(s,a) \kl_{\phi \mid \psi}(s,a) \geq 1 \Big\}, \label{eq:g_info_constraint}
\end{align}
and navigation constraints:
\begin{align}
\set{N}(\phi):= \Big\{&\eta \in \RR_{+}^{S \times A}: \forall s' \in \set{S}, \nonumber \\
&\sum_{a \in \set{A}(s')} \eta(s',a) = \sum_{s \in \set{S}}\sum_{a \in \set{A}(s)}p_\phi(s'|s,a) \eta(s,a)\Big\}. \label{eq:g_navi_constraint}
\end{align}
\end{theorem}
\proof The proof relies on two propositions which we introduce below. The first is an expression of regret as the sum of the number of times every sub-optimal state-action pair has been visited, first shown by \cite{burnetas1997optimal}:

\begin{proposition} \label{prop:BK1}
Let $\phi \in \Phi_D$. For any policy $\pi \in \Pi$ and any initial state $s_1 \in \set{S}$, we have, as $T \to \infty$:
\begin{align*}
R_T^\pi(s_1,\phi) = \sum_{(s,a) \in \set{S}\times \set{A}}\EE_{\phi|s_1}^\pi\left[N_T(s,a) \right]\left(g_\phi^\star - r_\phi(s,a) \right).
\end{align*} 
\end{proposition}

The second proposition is the result of a fundamental information inequality by \cite{garivier2018explore} and first adapted to MDPs by \cite{ok2018exploration}: 

\begin{proposition}
For all $\phi \in \Phi_D$, all $\psi \in \Delta(\phi)$, and any event $\set{E}$:
\begin{align} \nonumber
\sum_{(s,a) \notin C^\star_\phi}&\EE_{\phi|s_1}^\pi\left[N_T(s,a)\right]\kl_{\phi|\psi}(s,a) \\
&\geq \kl\left(\PP_{\phi|s_1}^\pi[\set{E}], \PP_{\psi|s_1}^\pi[\set{E}]\right).
    \label{eq:f-long}
\end{align}
\end{proposition}

Note that in the above sum, we do not consider $(s,a) \in C^\star_\phi$ since, by the definition of $\psi$, we have $\kl_{\phi|\psi}(s,a) = 0$ if $(s,a)\in C^\star_\phi$.

Now we select event $\set{E}$, taking advantage of the fact that $\pi$ is uniformly good. By definition of $\psi$, we have $\Pi(\phi) \cap \Pi(\psi) = \emptyset$, which implies that there exists $(s,a) \in C^\star_\phi \backslash C^\star_\psi$ such that, for all $\alpha > 0$,
\begin{align*}
\EE_{\psi|s_1}^\pi \left[N_T(s,a) \right] = o(T^{\alpha}),
\end{align*}
and
\begin{align*}
\EE_{\phi|s_1}^\pi\left[ N_T(s) - N_T(s,a) \right] = o(T^{\alpha}).
\end{align*}
We fix a state-action pair $(s,a)$ and note that $s$ is recurrent under an optimal policy for $\phi$, so there exists $\rho>0$ such that $\EE_{\phi|s_1}^\pi \left[ N_T(s) \right] = \rho T$ for large enough $T$.
We now define the event $\set{E}$ as:
\begin{align*}
\set{E}:=\left[N_T(s,a) \leq \rho T-\sqrt{T} \right].
\end{align*}
An application of Markov's inequality then yields
\begin{align*}
\PP^\pi_\phi\left[\set{E}\right] &= \PP^\pi_\phi\left[\rho T-N_T(s,a) \geq \sqrt{T}\right] \\
&\leq \frac{\EE_{\phi|s_1}^\pi\left[ N_T(s) - N_T(s,a) \right]}{\sqrt{T}},
\end{align*}
and
\begin{align*}
\PP^\pi_\psi\left[\set{E}^c\right] &= \PP^\pi_\psi\left[N_T(s,a) \geq \rho T - \sqrt{T}\right] \leq \frac{\EE_\psi^\pi\left[N_T(s,a) \right]}{\rho T-\sqrt{T}}. 
\end{align*}
It follows that $\PP_\phi^\pi\left[\set{E} \right] \to 0$ and $\PP_\psi^\pi\left[\set{E} \right] \to 1$, as $T \to \infty$. Therefore, we get:
\begin{align*}
\frac{\kl\left(\PP_{\phi|s_1}^\pi[\set{E}], \PP_{\psi|s_1}^\pi[\set{E}]\right)}{\log T} &\xrightarrow[T\to \infty]{}\frac{1}{\log T}\log \left( \frac{1}{\PP_{\psi|s_1}^\pi[\set{E}^c]} \right) \\
&\geq \frac{1}{\log T}\log \left( \frac{\rho T-\sqrt{T}}{\EE_\psi^\pi\left[N_T(s,a) \right]} \right),
\end{align*}
whose left-hand side converges to $1$ as $T \to \infty$ as a consequence of our choice of $(s,a)$. Substituting this result in \eqref{eq:f-long} yields:
\begin{align*}
    \liminf_{T\rightarrow \infty}\frac{1}{\log T} \sum_{s,a \notin \set{O}(s,\phi)}\EE_{\phi|s_1}^\pi\left[N_T(s,a)\right]\kl_{\phi|\psi}(s,a) \geq 1.
\end{align*}

This inequality, combined with Proposition \ref{prop:BK1}, yields the optimization problem along with the information constraints \eqref{eq:g_info_constraint}. The navigation constraints \eqref{eq:g_navi_constraint} follow directly from the deterministic transitions of problem $\phi$.
\ep

Theorem \ref{thm:g-r-lower-det} quantifies the minimal number of times a sub-optimal state-action pair $(s,a)$ must be visited by any uniformly good policy. This number scales as $\eta^\star(s,a)\log T$ where $\eta^\star(s,a)$ is the solution to the optimization problem \eqref{eq:g-r-lower-det} and thus implicitly defines the regret lower bound. The constraint sets $\set{I}(\phi)$ and $\set{N}(\phi)$ are respectively referred to as the \emph{information} and \emph{navigation} constraints. The former constrain the minimum number of visits on $(s,a)$ to discern between the problem $\phi$ and every confusing parameter $\psi$ while the later constraint the number of visits as a result of the transition dynamics.


\section{REGRET LOWER BOUNDS FOR DISJOINT CYCLES}

A notable limitation of Theorem \ref{thm:g-r-lower-det} is that the optimization problem \eqref{eq:g-r-obj-det} is infinite dimensional (as a consequence of the information constraint set $\set{I}(\phi)$. However, it can be simplified to a finite-dimensional one in the case of DMDPs where every cycle in $\set{C}$ is disjoint. First, we note that $\Delta(\phi) = \cup_{C\in \set{C}} \Delta (C;\phi)$, where we define $\Delta (C;\phi)$ as the set of bad problems such that the cycle $C$ is optimal, i.e.,
\begin{align*}
\Delta(C;\phi) = \Big\{&\psi \in \Phi_D :  \phi \ll \psi, \\
&(i) \ C^\star_\psi = C, \\
&(ii)\ \kl_{\phi \mid \psi}(s,a) = 0, \forall (s,a) \in C^\star_\phi  \Big\}.
\end{align*}
\textbf{Decoupling of information constraints.} Our aim is to reduce the dimension of the information constraints \eqref{eq:g_info_constraint}. To this end, we first decouple them by expressing the information constraint set as:
\begin{align*}
\set{I}(\phi) = \Big\{ &\eta \in \RR_{+}^{S\times A}: \forall C \neq C^\star_\phi, \\
\sum_{(s,a) \notin C_\phi^\star}& \eta(s,a) \kl_{\phi \mid \psi}(s,a) \geq 1, \ \forall \psi \in \Delta(C;\phi) \Big\},
\end{align*}
so that for each cycle $C$ there exists a set of constraints that must be satisfied by the solution of \eqref{eq:g-r-obj-det}. Every cycle can be associated to a maximally confusing parameter $\psi$, obtained by solving, for any $\eta(s,a)$ and a fixed $C$, the following optimization problem:
\begin{align} \label{eq:info_opt_prob}
\min_{\psi \in \Delta(C;\phi)} \sum_{(s,a) \notin C^\star_\phi} \eta(s,a)\kl_{\phi|\psi}(s,a).
\end{align}
Using the fact that all cycles C are disjoint, the solution to \eqref{eq:info_opt_prob} is obtained  by a $\psi$ that satisfies: $(i) \ g_\psi(C) = g^\star_\phi;~ (ii) \ \kl_{\phi|\psi}(s,a) = 0, \ \forall (s,a)\notin C$. This observation leads to the following constraint set:
\begin{align*}
\set{I}(\phi) = \Big\{ &\eta \in \RR_{+}^{S\times A}: \forall C \neq C^\star_\phi, \ \psi: g_\psi(C) = g^\star_\phi, \\
&\sum_{(s,a) \in C} \eta(s,a) \kl_{\phi \mid \psi}(s,a) \geq 1 \Big\} .
\end{align*}

\textbf{Intersection of navigation and information constraints.} We now examine the navigation constraints in \eqref{eq:g_navi_constraint}. Consider $\set{H}_T^\pi = (s_1,a_1,s_2,a_2,\dots, s_T,a_T)$, which is a trajectory induced by a policy $\pi\in \Pi$ starting at state $s_1$ up to time $T$. For any DMDP $\phi$, this trajectory is an open walk in a graph, and so it can be decomposed into a (possibly empty) set of cycles $C\in \set{C}$ and a finite path $P$ \cite{bang2008digraphs}. We denote by $N_T(C)$, the number of times a cycle $C$ appears in the decomposition of $H_T^\pi$. By definition, all cycles in $\phi$ are disjoint, and so, for all uniformly good policies, and for any state-action pair $(s,a) \in C \neq C^\star_\phi$, we have
\begin{align*}
\eta(s,a) &= \liminf_{T\to \infty} \frac{\EE_{\phi|s_1}^\pi [N_T(s,a)]}{\log T} \\
&= \liminf_{T\to \infty} \frac{\EE_{\phi | s_1}^{\pi} [N_T(C)]}{\log T} \\
& \ + \liminf_{T\to \infty} \frac{\EE_{\phi | s_1}^\pi\left[\sum_{i=1}^{|P|}\mathds{1}\{(s_i,a_i)=(s,a)\}\right] }{\log T} \\
&= \liminf_{T\to \infty}\frac{\EE_{\phi|s_1}^\pi [N_T(C)]}{\log T} := \eta(C).
\end{align*}
Thus, for uniformly good policies, the navigation constraints in \eqref{eq:g_navi_constraint} imply that every state-action pair in a cycle $C$ must be sampled at the same rate. In light of this observation, the information constraint set can be rewritten as:
\begin{align*}
\set{\set{I}(\phi)} = \Big\{&\eta \in \RR_{+}^{S \times A}: \forall C \neq C^\star_\phi, \ \psi: g_\psi(C) = g^\star_\phi, \\
&\eta(C)\sum_{(s,a) \in C}\kl_{\phi \mid \psi}(s,a) \geq 1 \Big\} .
\end{align*}
Now, recall that $q_\phi(s,a)$ belongs to the one-parameter exponential distribution family parameterized by the mean $r_\phi(s,a)$. Then, as explained in \cite{cappe2013kullback}, there exists a convex, twice differentiable function $b_{sa}(\phi)$ such that $r_\phi(s,a) = \dot{b}_{sa}(\phi)$. The Kullback Leibler divergence then becomes:
\begin{align*}
\kl(r_\phi(s,a),r_\psi(s,a)) = &b_{sa}(\psi) - b_{sa}(\phi) \nonumber \\
&-\dot{b}_{sa}(\phi)(r_\psi(s,a)-r_\phi(s,a)).
\end{align*}  
Thus we can rewrite \eqref{eq:info_opt_prob} as an optimization problem with respect to the mean reward vector $r_\psi$:
\begin{align}
\min_{r_\psi \in [0,1]^{|C|}}& \sum_{(s,a)\in C} \kl(r_\phi(s,a),r_\psi(s,a)) \label{eq:psi_objective} \\
\textnormal{subject to}& \ \sum_{(s,a)\in C} r_\psi(s,a) = |C|g^\star_\phi. \label{eq:mean_reward_constraints}
\end{align}

Finally, to simplify notation, we define the set of mean rewards that satisfy  \eqref{eq:mean_reward_constraints} as:
\begin{align*}
\set{R}(C;\phi):= \Big\{&r_\psi \in [0,1]^{|C|}: \sum_{(s,a)\in C} r_\psi(s,a) = |C|g^\star_\phi \Big\},
\end{align*}

The above simplifications lead to the following regret lower bound that is valid for any DMDP $\phi$ with disjoint simple cycles:
\begin{theorem} \label{thm:d-r-lower-det}
Let $\pi \in \Pi$ be a uniformly good policy. For all $\phi \in \Phi_D$ with disjoint cycles, and initial $s_1 \in \set{S}$ we have:
\begin{align} \label{eq:d-r-lower-det}
\liminf_{T \to \infty} \frac{R^\pi_T (s_1,\phi)}{\log T} \ge C(\phi), 
\end{align}
where $C(\phi)$ is the value of the optimization problem:
\begin{align}
\min_{\eta \in \set{I}(\phi)\cap \set{N}(\phi)} \sum_{(s,a)\in \set{S}\times \set{A}} \eta(s,a)(g^\star_\phi - r_\phi(s,a)), \label{eq:d-r-obj-det}
\end{align}
with information constraints:
\begin{align}
\set{I}(\phi)= \Big\{&\eta \in \RR_{+}^{S \times A}: \forall C \neq C^\star_\phi, \nonumber \\
\eta(C)&\min_{r_\psi \in \set{R}(C;\phi)} \sum_{(s,a) \in C} \kl(r_\phi(s,a), r_\psi(s,a)) \geq 1 \Big\}, \label{eq:d_info_constraint}
\end{align}
and navigation constraints:
\begin{align}
\set{N}(\phi)= \big\{&\eta \in \RR_{+}^{S \times A}: \forall C \neq C^\star_\phi, \nonumber\\
&\eta(s,a) = \eta(C), \ \forall (s,a) \in C \big\}. \label{eq:d_navi_constraint}
\end{align}
\end{theorem}

Theorem \ref{thm:d-r-lower-det} shares the same interpretation as Theorem \ref{thm:g-r-lower-det} with the notable difference that we now have two finite-dimensional nested convex optimization problems which can be solved efficiently. In order to determine the lower bound, we must first solve \eqref{eq:psi_objective} for each cycle $C$. Each solution corresponds to the most confusing parameter $\psi$ with the property that its associated cycle is optimal. Together they form $|C|$ constraints which, combined with the $|C|$ navigation constraints \eqref{eq:d_navi_constraint} comprise the set of constraints for the regret minimization problem \eqref{eq:d-r-obj-det}.


\section{EXAMPLES}

In this section we present two example applications that can be modeled as DMDPs with disjoint cycles and for which we can state the regret lower bound explicitly. We also highlight the relationship between DMDPs with disjoint cycles and the multi-armed bandit problem.

\subsection{Deterministic Line Search}

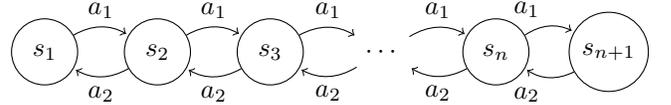
\begin{figure}[ht]
\centering
\begin{tikzpicture}[shorten >=1pt,node distance=1.5cm,on grid,auto]
    \node[state] (q_0) {$s_1$};
    \node[state] (q_1) [right=of q_0] {$s_2$};
    \node[state] (q_2) [right=of q_1] {$s_3$};
    \node[rectangle] (q_dots) [right=of q_2] {$\cdots$}; 
    \node[state] (q_3) [right=of q_dots] {{$s_{n}$}};    
    \node[state] (q_4) [right=of q_3] {$s_{n+1}$};

    \path[->]
    (q_0) edge [bend left] node {$a_1$} (q_1)
    (q_1) edge [bend left] node {$a_2$} (q_0)
    (q_1) edge [bend left] node {$a_1$} (q_2)
    (q_2) edge [bend left] node {$a_2$} (q_1)
    (q_2) edge [bend left] node {$a_1$} (q_dots)
    (q_dots) edge [bend left] node {$a_2$} (q_2)
    (q_dots) edge [bend left] node {$a_1$} (q_3)
    (q_3) edge [bend left] node {$a_2$} (q_dots) 
    (q_3) edge [bend left] node {$a_1$} (q_4)
    (q_4) edge [bend left] node {$a_2$} (q_3);
\end{tikzpicture}
\caption{Deterministic Line Search problem.}
\label{fig:Chain}
\end{figure}

For our first example, consider a DMDP $\phi \in \Phi_D$ with state space $\set{S} = \{s_1,\dots,s_n+1\}$ and action space $\set{A} = \{a_1,a_2\}$, such that $\set{A}(s_1) = {a_1}$, $\set{A}(s_{n+1}) = \{a_2\}$, and $\set{A}(s_i) = \{a_1,a_2\}, \forall i\in \{2,\dots,n \}$. The DMDP has deterministic transitions as illustrated in Figure \ref{fig:Chain}. 

As we can see, this DMDP consists of $n$ cycles $C_i=((s_i,a_1),(s_{i+1},a_2))$ for $i \in \{1, \dots, n \}$. For every such deterministic line search problem $\phi$, each cycle $C_i$ has gain
\begin{align*}
g_\phi(C_i) = \frac{r_\phi(s_i,a_1) + r_\phi(s_{i+1},a_2)}{2}.
\end{align*}

Denote by $C_j = C^\star_\phi$ the optimal cycle for $\phi$. We first express the set of bad parameters $\Delta(\phi)$ as a union over the sets $\Delta(C_i;\phi)$:
\begin{align*}
\Delta(C_i;\phi):= \{&\psi \in \Phi_D: \phi \ll \psi, \\
&i) \ g_\psi(C_i) = g^\star_\phi,  \\
&ii) \ \kl_{\phi|\psi}(s_j,a_1) + \kl_{\phi|\psi}(s_{j+1},a_2) = 0 \}.
\end{align*}

Then, the set of navigation constraints is simply:
\begin{align}
\set{N}(\phi):= \big\{&\eta \in \RR_{+}^{S \times A}: , \ \forall C_i \neq C_j, \nonumber \\
&\eta(s_i,a_1) = \eta(s_{i+1},a_2) = \eta(C_i)\big\}. \label{eq:line_navi_constraint}
\end{align}

The information constraints can be found by solving for every cycle $C_i$, the following convex optimization problem:
\begin{align*}
\min_{r_\psi \in [0,1]^2}&\kl(r_\phi(s_i,a_1), r_\psi(s_i,a_1)) \\
 &+\kl(r_\phi(s_i,a_1), r_\psi(s_{i+1},a_2)) \\
\textnormal{subject to}& \ r_\psi(s_i,a_1) + r_\psi(s_{i+1},a_2) = 2g^\star_\phi.
\end{align*}
This problem can be solved analytically, with its solution satisfying the pair of equations:
\begin{align*}
r^\star_\psi(s_j,a_1) &= g^\star_\phi + g_\phi(C_i) - r_\phi(s_{j+1},a_2), \\
r^\star_\psi(s_{j+1},a_2) &= g^\star_\phi + g_\phi(C_i) - r_\phi(s_{j},a_1).
\end{align*}
Using this solution, we can define the information number of cycle $C_i$ as
\begin{align*}
I(C_i):=& \kl\left(r_\phi(s_i,a_1),g^\star_\phi + g_\phi(s_i)-r_\phi(s_{i+1},a_2)\right) \\
&+\kl\left(r_\phi(s_{i+1},a_2),g^\star_\phi + g_\phi(s_i)-r_\phi(s_i,a_1)\right),
\end{align*}
and we attain the following information constraint set:
\begin{align}
\set{I}(\phi)= \big\{&\eta \in \RR_{+}^{S \times A}: \forall C_i \neq C_j, \nonumber \\
&\eta(C_i)I(C_i) \geq 1 \big\}, \label{eq:line_info_constraint}
\end{align}
In light of \eqref{eq:line_navi_constraint} and \eqref{eq:line_info_constraint}, we can rewrite the optimization problem \eqref{eq:d-r-lower-det} as:
\begin{align*}
\min_{\eta \geq 0}&\sum_{i=1}^n 2 \eta(C_i)\left(g_\phi^\star - g_\phi(C_i)\right), \\
\textnormal{subject to}& \ \eta(C_i)I(C_i) \geq 1, \ \forall C_i \neq C_j,
\end{align*}
with solution $\eta^\star(C_i) = 1/I(C_i)$. This leads to the explicit regret lower bound:

\begin{corollary} \label{cor:det_line}
Let $\pi \in \Pi$ be a uniformly good policy. For all deterministic line search problems $\phi$ and initial $s_1 \in \set{S}$ we have:
\begin{align*}
\liminf_{T\to \infty} \frac{R^\pi_T(\phi,s_1)}{\log T} \geq \sum_{i=1}^n \frac{2(g^\star_\phi - g_\phi(C_i))}{I(C_i)}.
\end{align*}
\end{corollary}

In contrast to Theorems \ref{thm:g-r-lower-det} and \ref{thm:d-r-lower-det}, the lower bound in Corollary \ref{cor:det_line} is explicit and analogous to \eqref{eq:lai-robbins} which corresponds to the problem-specific lower bound for the multi-armed bandit problem. Treating $g_\phi(C_i)$ as the reward of the cycle $C_i$ and the information number $I(C_i)$ as a similarity metric between the rewards of the cycle $C_i$ and those of the optimal cycle, we can interpret the deterministic line search problem as a case of multi-armed bandit where each cycle is an arm.

\subsection{State-dependent Rewards}
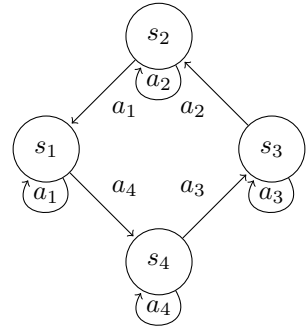
\begin{figure}[ht]
\centering
\begin{tikzpicture}[shorten >=1pt,node distance=1.5cm,on grid,auto]
    \node[state] (q_0) {$s_1$};
    \node[rectangle] (q_1) [right=of q_0] {};
    \node[state] (q_2) [right=of q_1] {$s_3$};
    \node[state] (q_3) [below=of q_1] {{$s_{4}$}};    
    \node[state] (q_4) [above=of q_1] {$s_{2}$};

    \path[->]
    (q_0) edge [loop above] node{$a_1$} (q_0)
    (q_2) edge [loop above] node{$a_3$} (q_2)
    (q_3) edge [loop above] node{$a_4$} (q_3)
    (q_4) edge [loop above] node{$a_2$} (q_4)
    (q_0) edge [] node {$a_4$} (q_3)
    (q_3) edge [] node {$a_3$} (q_2)
    (q_2) edge [] node {$a_2$} (q_4)
    (q_4) edge [] node {$a_1$} (q_0);
\end{tikzpicture}
\caption{Four state DMDP with state-dependent rewards.}
\label{fig:state-rewards}
\end{figure}

We now consider DMDPs with state-dependent rewards, i.e, DMDPs $\phi \in \Phi_D$ with reward distributions $q_\phi(s)$ having mean $r_\phi(s)$. Because of the deterministic structure, this is equivalent to having the same reward distribution for every state-action pair that leads to the same state. That is, for any two state-action pairs $(s_1,a_1)$, $(s_2,a_2)$ that satisfy $p_\phi(s'|s_1,a_1) = p_\phi(s'|s_2,a_2) = 1$, for a state $s'$, it holds that $q_\phi(\cdot |s_1,a_1) = q_\phi(\cdot |s_2,a_2) := q_\phi(\cdot|s')$. We further assume that it is always possible to stay in the same state, i.e., there exists $a \in \set{A}$ such that $p_{\phi}(s|s,a) = 1$. Actions that lead to the same state can be treated as identical, and thus the new action set becomes $\set{A} = \set{S}$. An example of such a problem is given in Figure in \ref{fig:state-rewards}.

For any state-dependent reward problem $\phi$, we define the optimal state as $s^\star_\phi = \argmax_{s \in \set{S}} r_\phi(s)$ and its associated optimal mean reward as $r_\phi^\star = r_\phi(s^\star_\phi)$. Let $N_T(s|s')$ be the number of times state $s$ has been chosen from state $s'$ up to time $T$ and let $N_T(s) = \sum_{s'\in \set{S}}N_T(s|s')$ be the total number of times $s$ has been visited. Similarly, we define $\eta(s|s') = \EE[N_T(s|s')]\log T$ and $\eta(s) = \EE[N_T(s)]\log T$. For every state $s$ we define its set of predecessors $\set{P}(s) := \{s' \in \set{S}: s \in \set{A}(s')\}$ and its set of descendants $\set{D}(s) := \{s' \in \set{S}: s' \in \set{A}(s)\}$.

We now show that in problems with state-dependent rewards, we recover the multi-armed-bandit setting, and our bound matches that of Lai and Robbins \cite{lai1985asymptotically}. This has two important implications. Firstly, that our lower bound is attainable (tight) and secondly, that it is possible to devise an optimal algorithm whose regret due to navigating a DMDP is sub-logarithmic (otherwise, there would be an added term on the regret due to navigation).

\begin{corollary} \label{cor:state_dependent}
Let $\pi \in \Pi$ be a uniformly good policy. For all DMDPs $\phi$ with state-dependent rewards and initial $s_1 \in \set{S}$ we have:
\begin{align}
\liminf_{T\to \infty} \frac{R^\pi_T(\phi,s_1)}{\log T} \geq \sum_{s \neq s^\star_\phi} \frac{r^\star_\phi - r_\phi(s)}{\kl(r_\phi(s),r^\star_\phi)} \label{eq:lai-robbins}.
\end{align}
\end{corollary}
\proof First observe that, as a consequence of our definitions above, we can rewrite \eqref{eq:g-r-obj-det}, \eqref{eq:g_info_constraint}, and \eqref{eq:g_navi_constraint} as follows:
\begin{align} \label{eq:so_obj}
\min_{\eta \geq 0}&\sum_{s \in \set{S}} \eta(s|s)(r^\star_\phi - r_\phi(s)) + \sum_{s'\neq s}\eta(s|s')(r^\star_\phi - r_\phi(s)) \\ 
\textnormal{subject to}& \ \sum_{s \neq s^\star_\phi} \eta(s|s) \kl_{\phi|\psi}(s) \geq 1, \ \forall \psi \in \Delta(\phi),\label{eq:so_info_constraint} \\ 
&\sum_{s' \in \set{P}(s)} \eta(s|s') = \sum_{s'\in \set{D}(s)}\eta(s'|s), \ \forall s \in \set{S} \label{eq:so_navi_constraint}
\end{align}

Where we use the fact that, since it is always possible to stay in the same state, we will have $C^\star_\phi = \{s^\star_\phi\}$ and thus $g^\star_\phi = r^\star_\phi$. As a consequence, we can restrict our attention to cycles consisting of a single state, so that the set of bad parameters becomes $\Delta(\phi) = \cup_{s \neq s^\star_\phi}\Delta(s;\phi)$, defined for every $s \in \set{S}$,
\begin{align*}
\Delta(s;\phi):= \{&\psi \in \Phi_D: \phi \ll \psi, \\
&i) \ r_\psi(s) \geq r^\star_\phi,  \\
&ii) \ \kl_{\phi|\psi}(s^\star_\phi) = 0 \}.
\end{align*}
In words, the set of bad parameters $\psi$ where the reward distribution at $s^\star_\phi$ is unchanged and a new state $s$ becomes optimal.

To recover the bound of Lai and Robbins \cite{lai1985asymptotically}, we must find for each state $s$ the associated bad problem $\psi$ that minimizes the left-hand side of the information constraints \eqref{eq:so_info_constraint}. This is achieved by choosing $\psi \in \Delta(s;\phi)$ such that $\kl_{\phi|\psi}(s') = 0, \ \forall s' \neq s$ and $r^\star_\psi(s) = r^\star_\phi$. Substituting this $\psi$ for every state and solving the optimization problem \eqref{eq:so_obj}, \eqref{eq:so_info_constraint} yields the solution:
\begin{align*}
\eta^\star(s) = \frac{1}{\kl(r_\phi(s),r^\star_\phi)}.
\end{align*}

It suffices to verify that this solution satisfies the navigation constraints \eqref{eq:so_navi_constraint}. Indeed, we have, for all $s \in \set{S}$,
\begin{align*}
\eta(s) = \sum_{s' \in \set{S}} \eta(s|s') = \sum_{s' \in P(s)} \eta(s|s') = \sum_{s'\in \set{D}(s)}\eta(s'|s)
\end{align*}

Where the first equality follows from the definition of $\eta(s)$, the second from the observation that $\eta(s|s') = 0$, $\forall s' \notin \set{P}(s)$, and the last equality from \eqref{eq:so_navi_constraint}.

Finally, since it is always possible to stay in the same state, we can let $\eta(s) = \eta(s|s)$ for every $s \in \set{S}$. Essentially, the exploration rates every state are decoupled, which complies with the solution $\eta^\star(s)$.
\ep


\section{CONCLUDING REMARKS}
For the case of communicating MDPs with deterministic transitions, we presented a general asymptotic regret lower bound that is problem specific and has the form of an infinite dimensional linear program. We focused on the case of DMDPs with disjoint cycles, and showed that this problem class has a decoupling property that allows us to simplify the regret lower bound by reducing it to two nested finite-dimensional linear programs. We exemplified this result by presenting the deterministic line search problem as well as DMDPs with state-dependent rewards. For these, we solved the associated optimization problems analytically and identified their similarities with the multi-armed bandit problem.

Notably, the fact that our lower bound matches that of Lai and Robbins \cite{lai1985asymptotically} points to the surprising and counter-intuitive result that we need not experience additional regret due to navigation. 


\bibliography{RL.bib}
\bibliographystyle{IEEEtran}

\end{document}